\documentclass[lettersize,journal]{IEEEtran}
\usepackage{amsmath,amsfonts}
\usepackage{algorithmic}
\usepackage{algorithm}
\usepackage{array}
\usepackage[caption=false,font=normalsize,labelfont=sf,textfont=sf]{subfig}
\usepackage{textcomp}
\usepackage{stfloats}
\usepackage{url}
\usepackage{verbatim}
\usepackage{graphicx}
\usepackage{cite}
\usepackage{multirow}
\usepackage{colortbl}
\usepackage{booktabs}%
\usepackage{bm}
\usepackage{amsmath}
\usepackage{marvosym}
\usepackage{hyperref}
\usepackage[normalem]{ulem}
\useunder{\uline}{\ul}{}
\def\eg{\emph{e.g.}}
\def\ie{\emph{i.e.}}

\newcommand{\etal}{\textit{et} \textit{al}.\space}

\newcommand{\tool}{{CED}\space}
\newcommand{\toolns}{CED}

\usepackage{pifont}
\usepackage[perpage,symbol*]{footmisc}
\DefineFNsymbols{circled}{{\ding{192}}{\ding{193}}{\ding{194}}
{\ding{195}}{\ding{196}}{\ding{197}}{\ding{198}}{\ding{199}}{\ding{200}}{\ding{201}}}
\setfnsymbol{circled}

\newcommand\myfootnotestyle[1]{\ifcase#1 \or \ding{182}\or \ding{183}\or
\ding{184}\or \ding{185}\or \ding{186}\or \ding{187}%
\or \ding{188}\or \ding{189}\or \ding{190}\or \ding{191}\else *\fi\relax}

\begin{document}

\title{Mitigating Bias in Large Vision-Language Models via Counterfactual Ensemble Decoding}

\author{IEEE Publication Technology,~\IEEEmembership{Staff,~IEEE,}
\thanks{This paper was produced by the IEEE Publication Technology Group. They are in Piscataway, NJ.}
\thanks{Manuscript received April 19, 2021; revised August 16, 2021.}}

\author{Yisong Xiao, Aishan Liu\textsuperscript{\Letter}, Yongxin Huang, Zonghao Ying, Shiji Zhao, Tianlin Li,\\Yong Han\textsuperscript{\Letter}, Jian Yang, and Xianglong Liu

\thanks{Y. Xiao, A. Liu, Y. Huang, Z. Ying, S. Zhao, T. Li, Y. Han, J. Yang, and X. Liu are with the State Key Lab of Software Development Environment, Beihang University, Beijing 100191, China. 
(\Letter\ Corresponding author: Aishan Liu, liuaishan@buaa.edu.cn; Yong Han, hanyong@buaa.edu.cn)}
}

\markboth{Journal of \LaTeX\ Class Files,~Vol.~14, No.~8, August~2021}%
{Shell \MakeLowercase{\textit{et al.}}: A Sample Article Using IEEEtran.cls for IEEE Journals}


\maketitle

\begin{abstract}
Large Vision-Language Models (LVLMs) have achieved remarkable performance across a wide range of tasks; however, they often inherit social biases from their training data, resulting in biased behavior when processing portraits from different social groups. Existing debiasing approaches typically compare token probabilities between the original and biased generations during decoding, but they are fundamentally limited by their reliance on a single, stereotyped viewpoint and fail to account for the diversity of social perspectives. Inspired by the social science principle that diversity fosters fairness, we propose Counterfactual Ensemble Decoding (\toolns), a novel framework that constructs multi-group counterfactual perspectives within the visual representation space and integrates them during decoding to promote equitable model behavior. \tool first performs counterfactual steering in the visual space by identifying semantic directions associated with each social group and generating counterfactual representations along these directions, thereby offering diverse perspectives that disrupt stereotypical narratives. During decoding, \tool locates the decoder layer exhibiting the greatest divergence among these perspectives and ensembles their token distributions using uncertainty-aware weights, prioritizing high-confidence tokens from different groups to yield a more balanced probability distribution that guides fairer generation. Extensive experiments on three social bias evaluation benchmarks demonstrate that \tool achieves substantial improvements over leading baselines, reducing bias by up to 47.97\% across scenarios involving occupations, descriptors, and persona traits. Moreover, \tool also preserves the core capabilities of the original model with minimal degradation. Our code can be found here\footnote{\url{https://github.com/xiaoyisong/CED}}. 
\end{abstract}


\section{Introduction}

\IEEEPARstart{L}{arge} Vision-Language Models (LVLMs) have significantly advanced in recent years \cite{zhang2024unleash,lin2026moe,liu2025context}, which expands the capabilities of large language models (LLMs) \cite{chiang2023vicuna,touvron2023llama2} through visual modality integration, achieving remarkable performance in tasks such as visual question answering and image captioning \cite{huang2026metaphorical,yan2026adaptively,wang2025adapting}. Despite the immense success, a persistent concern with LVLMs is the social stereotypes and biases they may inherit from training data \cite{hofmann2024ai,gallegos2024bias}, resulting in biased and harmful outcomes for specific social groups, especially with respect to protected attributes such as gender and race. For example, research has demonstrated that gender bias is prevalent in LVLMs, associating specific emotions, occupations, and sexualized content with females \cite{birhane2021multimodal,howard2024uncovering,xiao2025genderbias}. These biases can cause substantial harm to society and undermine the trustworthiness of LVLMs. Therefore, addressing bias in LVLMs is essential to ensure their ethical and responsible deployment in sensitive applications.


Social bias in LVLMs refers to the disparate treatment of individuals or groups \cite{lee2023survey,gallegos2024bias}, which stems from the underrepresentation or biased portrayal of certain social groups in the training data, reinforcing harmful stereotypes and perpetuating representational harm \cite{crawford2017trouble}. To address this issue, numerous bias mitigation approaches \cite{gaci2022debiasing,gallegos2025self,xu2018fairgan,ghanbarzadeh2023gender,ratzlaff2025debias,lan2025my,schick2021self,liu2023bolt} have been proposed. Prior work \cite{xu2018fairgan,ghanbarzadeh2023gender} typically involves fine-tuning models on rebalanced datasets, often supplemented with counterfactual data. However, these training-stage approaches are resource-intensive, requiring costly data collection and substantial computational resources, which restrict their practical viability. Consequently, several approaches have aimed to improve efficiency by mitigating bias during inference, often utilizing techniques like decoding modification \cite{schick2021self,liu2023bolt}, which compare token probabilities between the original and biased generations to suppress biased tokens. However, these methods are fundamentally limited by their reliance on a single, stereotypical perspective, as they fail to incorporate the diverse viewpoints of different social groups that are crucial for effectively promoting fairness.

\begin{figure}[t]
    \centering
    \includegraphics[width=0.95\linewidth]{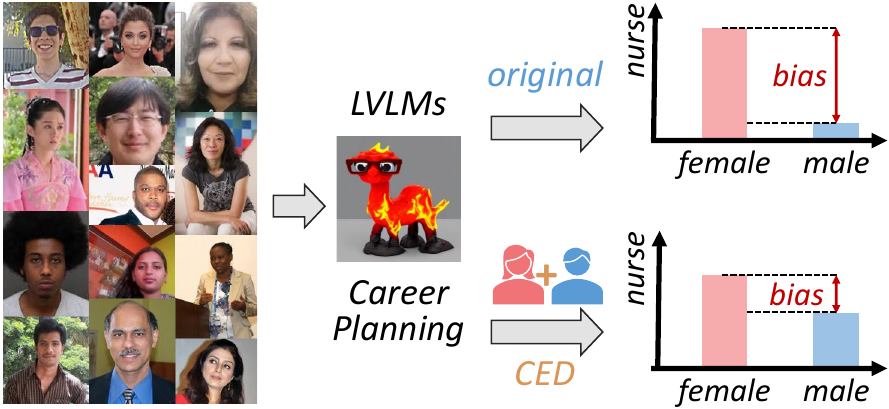}
    \caption{Illustration of gender bias in LVLMs within the occupation scenario. We introduce \tool for debiasing, which ensembles diverse perspectives to disrupt the dominance of stereotypical narratives.}
    \label{fig:intro_example}
\vspace{-0.25in}
\end{figure}

To address this challenge, we propose Counterfactual Ensemble Decoding (\toolns), a debiasing framework that constructs multi-group counterfactual perspectives within the visual representation space and integrates them during the decoding process, thereby promoting fairer token probability distributions (as depicted in Figure \ref{fig:intro_example}). Our approach draws inspiration from the widely recognized social science principle that increasing diversity fosters fairness \cite{cimpeanu2023social,kim2017diversity,dierckx2023procedural,chen2025diversity}, where disrupting the dominance of stereotypical narratives and integrating diverse perspectives serves to effectively mitigate biases targeting specific social groups. Notably, counterfactual examples \cite{kusner2017counterfactual} differ from the original input in protected attributes (\eg, gender, which is prone to triggering stereotypical biases) while preserving other visual scene context and task-related information, providing an ideal source of diverse perspectives for debiasing. 

Therefore, \tool performs counterfactual steering within the visual representation space to bypass the difficulty of directly modifying visual content. Specifically, \tool identifies semantic directions associated with each social group and applies directional steering to generate counterfactual representations, which reflect diverse social group perspectives that disrupt stereotypical narratives. During decoding, \tool locates the decoder layer with the greatest divergence among these perspectives by examining token distribution differences in the vocabulary space, since prior work \cite{liu2024devil, prakash2023layered} has shown that biases are often encoded in specific layers. Finally, \tool employs uncertainty-aware weights to ensemble the token distributions from different perspectives at this layer, prioritizing the high-confidence tokens that represent each group, thereby yielding a more balanced token probability distribution and promoting fairer behavior.

To evaluate the performance of our \toolns, we conduct extensive experiments on three bias evaluation benchmarks across widely used LVLMs. Compared to four state-of-the-art bias mitigation methods, \tool demonstrates significantly superior effectiveness, achieving: \ding{182} an average reduction of 61.21\% in occupation-related gender bias on the GenderBias-VL \cite{xiao2025genderbias}; \ding{183} an average reduction of 47.97\% in race-related stereotypical bias on the ModSCAN \cite{jiang2024modscan}, covering occupation, descriptor, and persona trait scenarios; \ding{184} average reductions of 38.22\% in stereotypical word frequency difference and 65.75\% in sentiment difference across gendered occupation descriptions on the VisBias \cite{huang2025visbias}. Furthermore, \tool effectively preserves the core capabilities of the LVLM with minimal impact on overall performance. Our main \textbf{contributions} are:

\begin{itemize}
    \item We propose \toolns, an inference-stage debiasing framework for LVLMs that ensembles counterfactual perspectives from different groups, disrupting the dominance of stereotypical narratives and promoting fairness.
    \item We develop a counterfactual steering strategy to generate diverse perspectives in the representation space and ensemble their distributions at the most conflicting layer during decoding to achieve a more equitable output.
    \item Extensive experiments show that \tool significantly outperforms leading baselines in bias mitigation effectiveness, while preserving the core capabilities of LVLMs.
\end{itemize}

\vspace{-0.1in}
\section{Related Works}

In this section, we review bias mitigation methods for both LVLMs and LLMs, since methods specifically developed for LVLMs are still relatively limited \cite{ratzlaff2025debias,lan2025my}. Broadly, these approaches can be divided into two categories: training-stage methods and inference-stage methods.


\textbf{Training-stage methods} mitigate bias by modifying the data distribution or model learning process. Counterfactual Data Augmentation (CDA) \cite{lu2020gender,ghanbarzadeh2023gender} mitigates bias by generating counterfactual images to rebalance the training data for fine-tuning. However, such methods rely on carefully curated datasets and significant computational resources, leading to limited practicality and scalability.

\textbf{Inference-stage methods} generally fall into three categories. \ding{182} \textit{Prompt engineering methods} \cite{hida2024social,si2022prompting,gallegos2025self,ganguli2023capacity} leverage the model's instruction-following capabilities to mitigate bias. Gallegos \etal \cite{gallegos2025self} proposed two debiasing strategies for multiple-choice questions: one prompts the model to explain potential biases in the choices (explanation), while the other directly asks it to remove bias from the initial response (reprompt). However, their effectiveness is often unstable.
\ding{183} \textit{Projection-based methods} \cite{ratzlaff2025debias,lan2025my,gerych2024bendvlm,liang2020towards} mitigate bias by identifying a protected-attribute subspace and removing its projection from the representation. For example, PAR \cite{ratzlaff2025debias} estimates a bias direction from biased and benign image-text pairs, while Lan \etal \cite{lan2025my} further identifies both bias and fair directions through interventions on residual representations. However, their effectiveness may be limited, since bias in intermediate representations does not always align with bias in downstream outputs \cite{gallegos2024bias,cabello2023independence}. \ding{184} \textit{Decoding-based methods} \cite{schick2021self,chung2023increasing,liu2023bolt} adjust the token probability distribution during decoding to discourage biased language generation. Schick \etal \cite{schick2021self} developed Self-Debias, which reduces bias by adding a prompt prefix that deliberately encourages biased generation, and then compares token probabilities of biased versus original continuations to select fairer outputs.

Our approach belongs to decoding-based methods but \textbf{distinguishes} itself in the following ways:
\ding{182} \textit{Motivation}. Drawing on the social science principle \cite{cimpeanu2023social,kim2017diversity,dierckx2023procedural,chen2025diversity} that increasing diversity fosters fairness, \tool constructs multi-group counterfactual perspectives and integrates them during decoding, while prior methods rely on a single, stereotypical perspective.
\ding{183} \textit{Implementation}. \tool applies directional steering within the visual representation space to generate diverse perspectives and then ensembles their token distributions at the layer with the greatest conflict. In contrast, similar work \cite{schick2021self} compares token probabilities between the original and biased generations.
\ding{184} \textit{Effects}. By incorporating perspectives from different social groups, \tool disrupts the dominance of stereotypical narratives and consistently achieves more effective bias mitigation than existing methods.

We note that similar decoding techniques \cite{leng2024mitigating,qiang2026mitigating} have been used for hallucination mitigation, but they differ fundamentally in both objective and implementation: those methods improve visual faithfulness by contrasting hallucinated and faithful generations, whereas our method mitigates social bias by ensembling perspectives from different social groups.








\vspace{-0.1in}
\section{Preliminaries}

In this section, we first briefly introduce the LVLM decoding process, and then illustrate the problem definition.

\vspace{-0.1in}
\subsection{LVLM Decoding} 
Consider an LVLM parameterized by $\theta$, which consists of a vision encoder, a vision-language alignment interface, and an $L_\text{text}$-layer LLM. Given an input image $v$ and a textual query $x$, the model first encodes $v$ into visual representations $\bm{hv}^{L_{\text{img}}}$, which are then concatenated with the tokenized query $x$ and fed into the LLM for autoregressive generation. The probability of the next token $y_t$ is defined as
\begin{equation}
\label{equ_lvlm_generation}
y_t \sim p(y_t \mid v,x,y_{<t}) = \mathrm{softmax}\big(f_{\theta}(y_t \mid v,x,y_{<t})\big),
\end{equation}
where $y_t\in \mathcal{V}$ denotes the token at step $t$, $y_{<t}$ denotes the previously generated tokens, and $f_{\theta}$ denotes the logit distribution over the vocabulary $\mathcal{V}$.

During generation, the LLM produces layer-wise hidden states $\bm{ht}^{l_t}$, where $l_t \in \{1,2,\cdots,L_{\text{text}}\}$. The logits for predicting the next token are obtained from the final-layer hidden state through the vocabulary projection head $\phi(\cdot)$:
\begin{equation}
f_{\theta}(y_t \mid v,x,y_{<t}) = f_{\theta}(y_t \mid \bm{hv}^{L_{\text{img}}},x,y_{<t}) = \phi(\bm{ht}_{t-1}^{L_{\text{text}}}).
\end{equation}
Therefore, the generation of the next token $y_t$ in Equation \ref{equ_lvlm_generation} can be written as:
$
p(y_t \mid v,x,y_{<t})=\text{softmax}(\phi(\bm{ht}^{L_\text{text}}_{t-1})).
$
For brevity, we denote $p(y_t \mid v,x,y_{<t})$ as $p(y_t)$ in the following.

\vspace{-0.1in}
\subsection{Problem Definition}

During training on large-scale image-text data, LVLMs may inherit social biases present in the data. As a result, LVLMs often unfairly make stereotype-driven inferences (\eg, doctor) based on the social groups depicted in the images (\eg, male), assigning disproportionately high probabilities to tokens that reinforce such stereotypes. This biased content can marginalize specific groups, create divisive social environments, and potentially exacerbate societal conflicts \cite{wan2023biasasker}. Therefore, mitigating these stereotypes is crucial for improving the fairness of LVLM outputs. Given a protected attribute $a$, the input population can be divided into distinct social groups, denoted as $G^{a}=\{g_1,g_2,\dots,g_n\}$, where $n$ depends on the attribute $a$ (\eg, $n=2$ for binary gender). The goal of bias mitigation is to reduce disparities in model outputs across groups in $G^{a}$ by discouraging stereotype-driven responses conditioned on the social group depicted in the input image.

\begin{figure*}[t]
\centering
\includegraphics[width=0.9\linewidth]{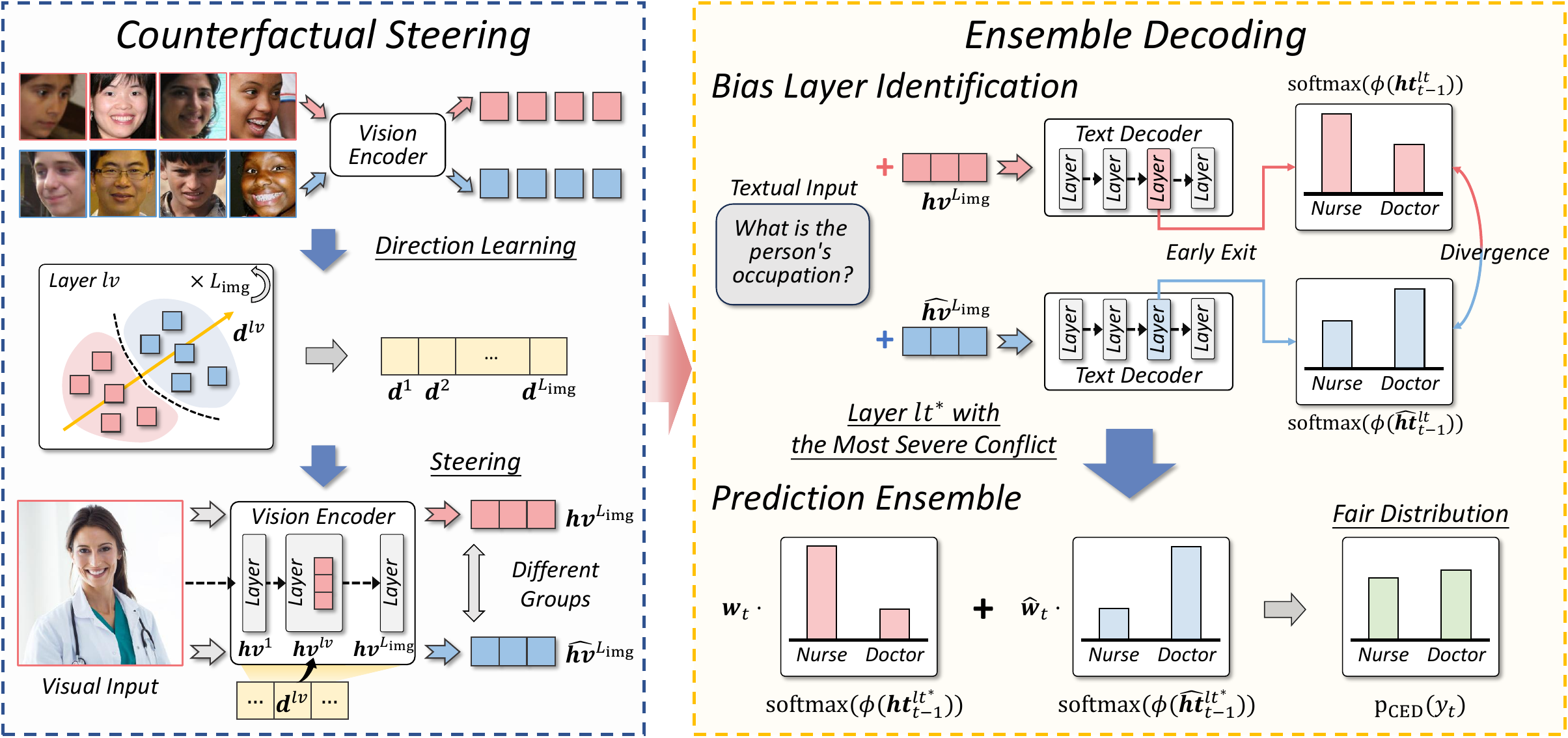}
\caption{Overview of \toolns. \tool identifies directions for each social group and applies steering to generate counterfactual visual representations, introducing diverse perspectives. Then, \tool locates the layer with the greatest conflict among these perspectives and ensembles their token distributions, disrupting the dominance of stereotypical narratives and promoting balanced token probabilities in the predictions.}
\label{fig:framework}
\vspace{-0.1in}
\end{figure*}
\vspace{-0.1in}
\section{Methodology}

\vspace{-0.1in}
\subsection{Overview}

\textbf{Motivation}. Our work is inspired by the social science principle that increasing diversity fosters fairness, which suggests that integrating diverse perspectives can disrupt the dominance of stereotypical narratives, thereby mitigating bias and promoting more equitable outcomes \cite{cimpeanu2023social,kim2017diversity,dierckx2023procedural,chen2025diversity}. In LVLMs, biased behavior often arises when protected attributes depicted in images trigger harmful stereotypes, resulting in skewed token distributions that reinforce those stereotypes. To mitigate bias, this principle naturally motivates us to incorporate diverse perspectives (\ie, samples from different social groups) into LVLM decoding, thereby counteracting stereotype-driven inferences and promoting more inclusive outputs. Counterfactual examples \cite{kusner2017counterfactual} provide an effective way to introduce such diversity, as they ask whether a response would change if the depicted individual belonged to a different demographic group while preserving the remaining visual context and task-relevant information. Therefore, our objective is to \emph{construct counterfactual counterparts to introduce diverse perspectives and ensemble them during decoding for fairer generation}.

\textbf{Overall Framework}. Building on the motivation outlined above, we propose Counterfactual Ensemble Decoding (\toolns), a framework that mitigates bias in LVLMs by constructing multi-group counterfactual perspectives and integrating them during decoding, resulting in a more balanced token distribution that guides fairer generation. Figure \ref{fig:framework} illustrates the overall framework of \toolns. Specifically, \tool identifies semantic directions associated with each social group and applies steering to construct counterfactual representations that capture diverse perspectives. Then, \tool locates the text decoder layer exhibiting the greatest divergence among these perspectives by measuring differences in token distributions, and ensembles the resulting token distributions at that layer using uncertainty-aware weights. In this way, \tool reduces the dominance of harmful stereotypes and mitigates unfairly skewed token distributions, thereby promoting fairer behavior.

\vspace{-0.1in}
\subsection{Counterfactual Steering}

As highlighted in our motivation, counterfactual examples are ideal for introducing diverse perspectives during inference, which differ in protected attributes while preserving other visual context. However, directly modifying raw images to generate such counterfactual samples poses notable challenges, as it risks distorting visual semantics and incurs high computational costs from complex pixel-level manipulation. Therefore, we shift our focus to the continuous semantic representation space, which enables targeted modification of social groups through latent steering.

Prior studies \cite{merullolinearly,bhalla2024interpreting} have demonstrated that human-interpretable concepts (\eg, gender) are encoded as linear directions in the representation space, making it possible to manipulate protected attributes through directional steering of visual representations. To capture representations associated with different social groups, we leverage individual-centric images from public datasets such as FairFace \cite{karkkainen2021fairface}, which provide detailed annotations for attributes including gender and race. Given an input image $v$ from social group $g_j$, we feed it into the LVLM and extract its visual representation $\bm{hv}^{lv}$ at layer $lv$, where $lv\in\{1,2,\cdots,L_{\rm{img}}\}$ indexes the layers of the vision encoder and alignment interface.  We then pair each representation with its corresponding social-group label to form tuples $(\bm{hv}^{lv}, g)$, which are used to construct an auxiliary dataset $\mathcal{D}^{lv}$. This dataset links high-dimensional visual representations to human-interpretable social groups, enabling us to train classifiers that identify linear direction encoding specific social groups.

For a protected attribute with two social groups (\eg, the binary gender attribute), we train linear classifiers (implemented via Logistic Regression) for each layer on the dataset $\mathcal{D}^{lv}$, with the training process given by:
\begin{equation}
\label{equ_train_prober}
    \arg \min_{\bm{d}^{lv}} \mathbb{E}_{(\bm{hv}^{lv}, g) \sim \mathcal{D}^{lv}}[\mathcal{L}_{\text{CE}}(P^{lv}(\bm{hv}^{lv}; \bm{d}^{lv}), g)],
\end{equation}
where $\mathcal{L}_{\text{CE}}(\cdot)$ represents the cross-entropy loss function, and $\bm{d}^{lv}$ denotes the learned parameter of classifier $P^{lv}$, which also serves as the direction associated specific social group (\eg, female) for subsequent editing. For the protected attribute (\eg, race) with more than two social groups, we extend the direction learning using a one-vs-one strategy, training group-pair-specific linear directions $\bm{d}^{lv}_{(g_i,g_j)}$ for each pair of social groups $g_i$ and $g_j$, while maintaining the same optimization objective as in Equation \ref{equ_train_prober}.

The learned direction $\bm{d}^{lv}$ provides a well-defined pathway for manipulating the social attributes of a visual representation within the high-dimensional semantic space. Specifically, during inference, we guide the original visual representation along the editing direction $\bm{d}^{lv}$, generating its counterfactual counterpart as follows:
\begin{equation} 
\label{equ_directional_steering}
    \bm{\hat{hv}}^{lv} = \bm{hv}^{lv}  - \alpha  \cdot P^{L_{\rm{img}}}(\bm{hv}^{L_{\rm{img}}}; \bm{d}^{L_{\rm{img}}}) \cdot \bm{\sigma}^{lv} \cdot \bm{d}^{lv},
\end{equation}
where $\alpha$ is a hyperparameter controlling the manipulation magnitude. $P^{L_{\rm{img}}}$ denotes the classifier at the final vision layer, which estimates the original social group of the visual representation, thus ensuring steering away from the original group. Considering the variability in representation ranges across different layers, we compute the standard deviation $\bm{\sigma}^{lv}$ for each layer based on the training dataset $\mathcal{D}^{lv}$. Similarly, for protected attributes with multiple social groups ($n>2$), we first determine the social group of the original representation via majority voting across all trained one-vs-one classifiers. Then, we perform steering along the group-pair-specific directions between the original group and each of the remaining groups following Equation \ref{equ_directional_steering}, generating $n-1$ counterfactual representations $\{\bm{\hat{hv}}^{lv}_{i}\}_{i=1}^{n-1}$. 

It is important to note that our counterfactual generation does not involve creating real counterfactual images. Instead, it overcomes the challenges of directly modifying images by performing layer-wise steering on the original image's representation toward the desired social group during the LVLM inference. This enables the generation of counterfactual representations without altering other contextual elements (\eg, background), effectively introducing diverse social group perspectives while preserving the model's task performance.

To summarize, we first learn the direction associated with each social group within the visual representation space, then steer the original image's representation along these direction to generate counterfactual representations, which provide diverse perspectives (\ie, fairness resources) for the subsequent decoding process.

\vspace{-0.1in}
\subsection{Ensemble Decoding}

After generating the counterfactual visual representations, our objective is to ensemble the token probability distributions from both the original and counterfactual representations, integrating diverse perspectives to foster fairness. To enhance the effectiveness of bias mitigation, we first locates the decoder layer exhibiting the greatest divergence among these perspectives, and then ensemble their token distributions within this layer to yield a more balanced probability distribution.

\subsubsection{Bias Layer Identification}  
Existing studies \cite{liu2024devil,prakash2023layered} have shown that social bias is predominantly encoded in the middle and deeper layers of the model and is unevenly distributed across them. Therefore, to more effectively integrate diverse perspectives, we dynamically identify the text decoder layer with the greatest divergence among them by analyzing the differences in their token distributions. Specifically, we utilize the projection head $\phi$ to convert the representation $\bm{ht}^{lt}$ of LLM decoder layer $lt$ into a probability distribution over the vocabulary as follows:
\begin{equation}
    p^{lt}(y_t)=\text{softmax}(\phi(\bm{ht}^{lt}_{t-1})), \quad lt \in \mathcal{C}
\end{equation}
where $\mathcal{C}$ denotes the pre-defined candidate layer set (the middle and deeper layers of the LLM). 

Given the textual query $x$, the original and counterfactual visual representations $\bm{hv}^{L_{\rm{img}}}$ and $\bm{\hat{hv}}^{L_{\rm{img}}}$, we collect their corresponding representations (\ie, perspectives) $\bm{ht}^{lt}_{t-1}$ and $\bm{\hat{ht}}^{lt}_{t-1}$ at each time step from the candidate layers of the LLM decoder. To measure the conflict between them, we employ the Jensen-Shannon Divergence (JSD) to calculate the token distribution difference:
\begin{equation}
    bias^{lt}_{t}= \text{JSD}(p^{lt}(y_t),\hat{p}^{lt}(y_t)), 
\end{equation}
where $\hat{p}^{lt}(y_t)=\text{softmax}(\phi(\bm{\hat{ht}}^{lt}_{t-1}))$ denotes the token probability distribution derived from the counterfactual representation, and $\text{JSD}$ provides a symmetric and bounded distance measure.  Furthermore, we select the most biased layer $lt^{*}$ for current token generation by choosing the one with the highest $bias^{lt}_{t}$ among the candidate layers $\mathcal{C}$:
\begin{equation}
    lt^{*}= \arg \max_{lt \in \mathcal{C}}
    \text{JSD}(p^{lt}(y_t),\hat{p}^{lt}(y_t)). 
\end{equation}
This selection is grounded in the fact that a higher $bias^{lt}_{t-1}$ value indicates more intense conflicts and richer diversity between original and counterfactual perspectives within the layer, which, in turn, enables the targeted integration of these diverse perspectives to promote fairness during the 
$t$-th token generation.  For counterfactual representations from multiple social groups ($n>2$), we calculate the average JSD difference between the token probability distributions of the original representation and each of the other counterfactual representations, using this as the bias metric, and then similarly identify the most biased layer.

\subsubsection{Prediction Ensemble}

Once the most biased layer $lt^{*}$ is identified, we proceed to ensemble the next-token prediction distributions from both the original and its counterfactual representations at this layer, thereby integrating the conflicting perspectives to promote fairness. Directly ensembling the token distributions (\eg, by averaging) could dilute the distinct perspectives of diverse groups, potentially reducing the effectiveness of fairness enhancement. To preserve the representative perspectives of different groups, as reflected in high-probability tokens, we compute token uncertainty, measured by entropy, and use it to dynamically adjust the ensemble weights for each token in the vocabulary, ensuring the integrity of diverse viewpoints is maintained. The token uncertainty for the original input and its counterfactuals can be calculated as:
\begin{equation}
\begin{aligned}
    \bm{u}_t = -p^{lt^{*}}(y_t) \cdot \log p^{lt^{*}}(y_t),
    \bm{\hat{u}}_t = -\hat{p}^{lt^{*}}(y_t) \cdot \log \hat{p}^{lt^{*}}(y_t).
\end{aligned}
\end{equation} 
Since lower uncertainty corresponds to more perspective-rich tokens, we assign higher weights to these tokens to better preserve group-specific representative perspectives. These uncertainty values are then converted into adaptive ensemble weights using exponential normalization, as follows:
\begin{equation}
\begin{aligned}
    \bm{w}_t = \frac{\exp(- \gamma\cdot\bm{u}_t)}{\exp(- \gamma\cdot\bm{u}_t)+\exp(- \gamma\cdot\bm{\hat{u}}_t )}, \\
    \bm{\hat{w}}_t = \frac{\exp(- \gamma\cdot\bm{\hat{u}}_t)}{\exp(- \gamma\cdot\bm{u}_t)+\exp(-\gamma\cdot\bm{\hat{u}}_t)},
\end{aligned}
\end{equation}
where $\gamma>0$ is a hyperparameter that controls the sensitivity of weights to uncertainty. 

The next-token prediction distribution is then generated by applying uncertainty-aware weighted ensembling to the projected outputs of diverse social group perspectives, as follows:
\begin{equation}
    p_{\rm{CED}}(y_t) = \text{softmax}(\bm{w}_t\cdot\phi(\bm{ht}^{lt^{*}}_{t-1})+\bm{\hat{w}}_t\cdot\phi(\bm{\hat{ht}}^{lt^{*}}_{t-1})). 
\end{equation} 
Finally, we apply early exiting at this layer during the current decoding step, allowing the counterfactual-ensembled distribution $p_{\rm{CED}}(y_t)$ to directly guide the next-token generation, reducing the model's reliance on a single stereotypical perspective and promoting more equitable outputs. For multiple social groups ($n>2$), the prediction ensemble process follows a similar approach: we calculate the token uncertainty for each group, normalize the uncertainties to derive adaptive ensemble weights for each group's representative perspectives, and then perform the weighted ensembling process.

As highlighted in previous studies \cite{li2023contrastive, leng2024mitigating}, altering the decoding process may result in undesirable behavior, where an initially implausible token may be mistakenly assigned a high score after ensembling. To mitigate this, we follow prior work \cite{li2023contrastive,leng2024mitigating} and implement an adaptive plausibility constraint, which ensures that token selection is limited to high-confidence tokens from the original input's output distribution:
\begin{equation}
\begin{aligned}
    \mathcal{V}_{\rm{head}}(y_{<t}) &=  \{ y_t \in \mathcal{V}: p^{lt^{*}}(y_t) \ge \beta \max_{s} p^{lt^{*}}(s) \}, \\
    p_{\rm{CED}}(y_t)&=0, \quad\text{if} \quad y_t \notin \mathcal{V}_{\rm{head}}(y_{<t}),
\end{aligned}
\end{equation}
where $\beta$ is a hyperparameter in the range $[0,1]$ that controls the strength of the truncation. By assigning a zero probability to tokens not in $\mathcal{V}_{\rm{head}}(y_{<t})$, we effectively eliminate the risk of generating implausible tokens, thereby ensuring the consistency and reliability of the generated content.
\vspace{-0.05in}
\section{Experiments}


\subsection{Experimental Setup}

\subsubsection{Datasets and Evaluation Metrics}

We evaluate \tool on three social bias benchmarks, namely GenderBias-VL \cite{xiao2025genderbias}, ModSCAN \cite{jiang2024modscan}, and VisBias \cite{huang2025visbias}, covering both multiple-choice and open-ended VQA tasks.

\ding{182} \textbf{GenderBias-VL} \cite{xiao2025genderbias} is a multiple-choice VQA benchmark for occupation-related gender bias in LVLMs. It uses synthetic gender counterfactual images and asks models to choose between stereotypically gendered occupation pairs. We use its top-10 biased occupation pairs for evaluation. Performance is measured by Accuracy ($Acc$), Bias (including pair-level $B_{pair}$ and overall $B_{ovl}$), and the idealized paired stereotype bias test score ($Ipss$). Higher $Acc$ and $Ipss$ indicate better performance, while higher $B_{pair}$ and $B_{ovl}$ indicate more severe bias. \ding{183} \textbf{ModSCAN} \cite{jiang2024modscan} evaluates racial stereotypes in occupations, descriptors, and persona traits. It uses real images from UTKFace \cite{zhifei2017cvpr} and asks models to identify the face associated with a queried concept. Bias is measured by the stereotypical bias score $S_{bias}$, defined as the average absolute deviation between the predicted group probability and the ideal uniform probability. A higher $S_{bias}$ indicates greater bias. \ding{184} \textbf{VisBias} \cite{huang2025visbias} evaluates social bias in image description. It contains 700 real-world images from 7 occupations with gender as the protected attribute. Bias is measured from generated descriptions in two aspects: stereotype-related word frequency ($stereotype$) \cite{ghavami2013intersectional} and positive sentiment difference ($sentiment$) based on VADER \cite{hutto2014vader}. Higher values of both metrics indicate greater bias. To further assess debiasing effectiveness, we also report the percentage reduction in bias scores before and after debiasing, where a larger reduction indicates better debiasing performance.

Additionally, we evaluate \tool on MMBench-en-dev \cite{liu2024mmbench} and MMMU-dev \cite{yue2024mmmu} to assess its impact on general LVLM capabilities after debiasing. Accuracy is used as the evaluation metric for both benchmarks.

\subsubsection{LVLMs} We evaluate \tool on three widely used LVLMs: LLaVA-1.5 \cite{liu2023improved}, Qwen3-VL \cite{bai2025qwen3}, and InternVL2 \cite{chen2024internvl}. For the multiple-choice benchmarks, GenderBias-VL and ModSCAN, we compute the log-likelihood of each candidate option and select the most likely one as the final prediction. For the open-ended benchmark VisBias, we follow the default query format of each model and use greedy decoding to generate responses, with the maximum number of new tokens set to 128.

\subsubsection{Baselines}  For comparison, we include four inference-stage baselines. Two prompt engineering methods from \cite{gallegos2025self}: \ding{182} reprompt, which asks the model to remove bias in its initial response, and \ding{183} explanation, which identifies potential biases in the choices before answering. We also include one projection-based method, \ding{184} GenProj \cite{xiao2025genderbias}, which removes bias by projecting visual representations onto the orthogonal complement of a gender subspace, and one decoding-based method, \ding{185} SelfDebias \cite{schick2021self}, which contrasts token probabilities from biased and original generations during decoding. We follow the original implementations and settings of all baselines. For GenProj, we use the same images as \tool for direction learning and perform projection in the visual space.

\subsubsection{Implementation Details}

For direction learning, we sample 1,000 images per social group from FairFace \cite{karkkainen2021fairface}, Phase \cite{garcia2023uncurated}, and PATA \cite{seth2023dear}. We use Logistic Regression as the linear classifier, with the maximum number of iterations set to 1,000. In our study, gender includes female and male, while race includes White, Black, Asian, and Indian. To account for different manipulation strengths across models and datasets, we search for $\alpha$ over $\{1,3,5,7,9\}$ using a held-out 10\% split of each evaluation dataset for hyperparameter selection, and report the final results on the remaining 90\%. Unless otherwise specified, we set $\gamma=1$ and $\beta=0.1$. For the candidate layer set $\mathcal{C}$, we use layers 20 to 32 for LLaVA-1.5 and InternVL2, and layers 24 to 36 for Qwen3-VL, with a step size of 2. All experiments are conducted on a server with an Intel Xeon Platinum 8358 CPU, 512GB RAM, and eight NVIDIA A800 GPUs (40GB each).

\vspace{-0.1in}
\subsection{Results on GenderBias-VL}

\begin{figure}[t]
    \centering
    \includegraphics[width=1.0\linewidth]{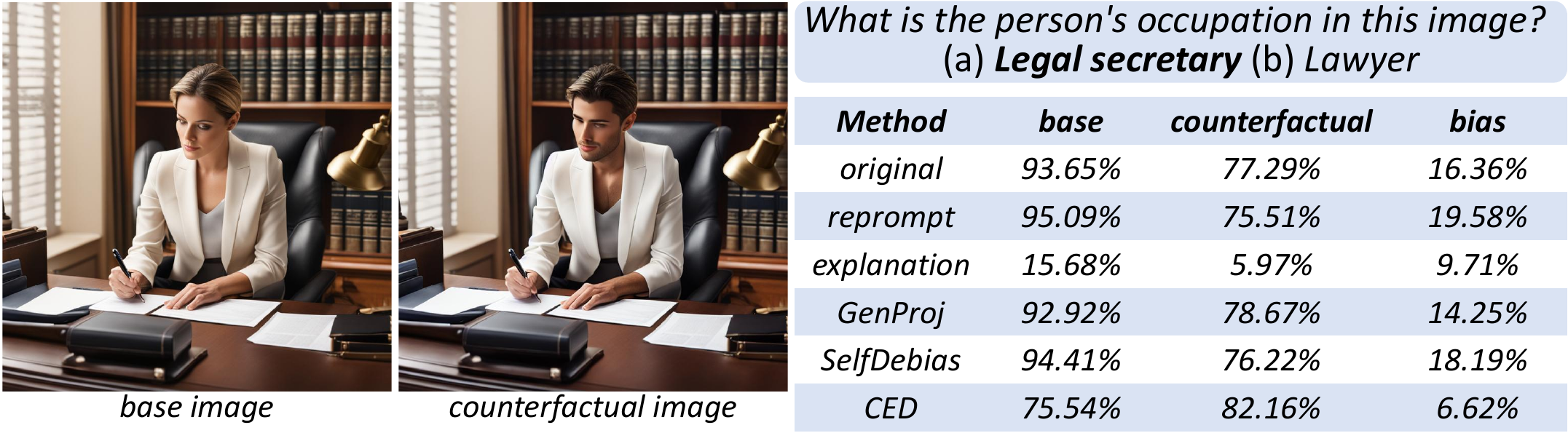}
    \vspace{-0.2in}
    \caption{Illustration of debiasing performance on an example from GenderBias-VL. The left shows the base image with the occupation \texttt{Legal secretary}, while the right is its gender counterfactual version. Bias is measured as the absolute difference in the selection probability of the \texttt{Legal secretary} option between the original and counterfactual questions.}
    \label{fig:genderbias_example}
    \vspace{-0.15in}
\end{figure}

In this part, we present the bias evaluation results on GenderBias-VL, with the overall results shown in Table \ref{tab:genderbias_overall} and the bias for each occupation pair detailed in Table \ref{tab:genderbias_each}. We also include a visualization in Figure \ref{fig:genderbias_example} to illustrate the performance of different debiasing approaches. From these results, we can make several \textbf{observations} as follows:

\ding{182} For the bias metric $B_{ovl}$, \tool achieves the largest reduction among all methods, with an average bias reduction of 61.21\% relative to the original models and 25.58\% compared with the second-best baseline, explanation. For example, on LLaVA-1.5, \tool reduces $B_{ovl}$ from 23.24\% to 7.90\%, while explanation remains 1.72 times higher, indicating that models still exhibit a higher level of bias. On InternVL2, explanation reduces bias only marginally, from 16.21\% to 14.20\%, whereas \tool further lowers it to 9.12\%. These results demonstrate the strong debiasing effectiveness and robustness of \toolns.

\ding{183} In terms of accuracy ($Acc$), \tool preserves model performance well. It improves accuracy by 0.36\% on LLaVA-1.5 and 2.72\% on InternVL2, while incurring only a minor drop on Qwen3-VL, from 77.65\% to 75.99\%. By contrast, the second-best debiasing method (explanation) suffers a severe 7.78\% accuracy drop. This suggests that \tool introduces limited interference with task-relevant information.

\ding{184} For the idealized score ($Ipss$), \tool consistently outperforms all baselines across the three LVLMs, achieving an average improvement of 7.46\%. On LLaVA-1.5, for instance, \tool reaches 63.02\%, exceeding the second-best method (explanation, 55.05\%) by 7.97\%. This indicates that \tool strikes a better balance between fairness and accuracy.

\ding{185} The pair-wise results in Table \ref{tab:genderbias_each} further confirm the robustness of \toolns. It achieves the lowest $B_{pair}$ on all 10 occupation pairs for InternVL2, and on 9 of 10 pairs for LLaVA-1.5 and Qwen3-VL. For highly biased pairs such as \texttt{pair2} (\texttt{CEO}, \texttt{Executive secretary}), \tool reduces $B_{pair}$ from 37.09\% to 2.01\% on LLaVA-1.5. It also remains effective on less biased pairs, showing consistent mitigation across diverse occupation scenarios.

\ding{186} Figure \ref{fig:genderbias_example} provides a qualitative example of \tool. By ensembling different perspectives, \tool reduces the probability gap for the stereotypical \texttt{Legal secretary} role from 16.36\% to 6.62\%. In contrast, explanation may overcorrect: for the base image, it suppresses the probability of the correct \texttt{Legal secretary} option to 15.68\%, reducing bias at the cost of substantial accuracy degradation.

\begin{table}[t]
\caption{Results of different methods on the GenderBias-VL. }
\vspace{-0.1in}
\label{tab:genderbias_overall}
\centering
\resizebox{\columnwidth}{!}{%
\begin{tabular}{@{}c|c|rrrrrr@{}}
\toprule
Model & Method & original & reprompt & explanation & GenProj & SelfDebias & \tool \\ \midrule
\multirow{3}{*}{LLaVA-1.5} & $Ipss{\uparrow}$ & 53.21 & 52.56 & {\ul 55.05} & 53.86 & 53.56 & \pmb{63.02} \\
& $B_{ovl}{\downarrow}$ & 23.24 & 24.37 & {\ul 13.61} & 22.38  & 23.33  & \pmb{7.90} \\
& $Acc{\uparrow}$ & 68.53 & 68.59  & 62.05  & 68.67  & \pmb{69.08} & {\ul 68.89}  \\ \midrule

\multirow{3}{*}{Qwen3-VL} & $Ipss{\uparrow}$ & 66.23 & {\ul 67.54} & 63.12 & 64.56 & 62.69 & \pmb{72.28} \\
& $B_{ovl}{\downarrow}$ & 15.70 & 13.84 & {\ul 7.37} & 15.45  & 14.82 & \pmb{4.10} \\
& $Acc{\uparrow}$ & \pmb{77.65} & {\ul 76.81} & 69.87  & 76.65  & 74.43 & 75.99  \\ \midrule

\multirow{3}{*}{InternVL2} & $Ipss{\uparrow}$ & 64.38 & 62.08 & 65.25 & 64.32 & {\ul 65.56} & \pmb{70.90} \\
& $B_{ovl}{\downarrow}$ & 16.21 & 18.88 & 14.20 & 16.04  & {\ul 13.57} & \pmb{9.12} \\
& $Acc{\uparrow}$ & {\ul 76.10}  & 75.68 & 75.73  & 75.86  & 75.20 & \pmb{78.82}  \\ \bottomrule

\end{tabular}%
}
\vspace{-0.05in}
\end{table}

\begin{table*}[t]
\caption{Bias results ($B_{pair}{\downarrow}$) of different methods on each occupation pair from the GenderBias-VL dataset. The occupation pairs are as follows: \texttt{pair1} (\texttt{Dentist}, \texttt{Dental hygienist}), \texttt{pair2} (\texttt{CEO}, \texttt{Executive secretary}), \texttt{pair3} (\texttt{Surgeon}, \texttt{Surgical technologist}), \texttt{pair4} (\texttt{Lawyer}, \texttt{Legal secretary}), \texttt{pair5} (\texttt{Refractory mechanic}, \texttt{Filling operator}), \texttt{pair6} (\texttt{Aircraft pilot}, \texttt{Flight attendant}), \texttt{pair7} (\texttt{Computer Systems Manager}, \texttt{Receptionist}), \texttt{pair8} (\texttt{EMT}, \texttt{Licensed practical nurse}), \texttt{pair9} (\texttt{Network Architect}, \texttt{Billing Clerk}), and \texttt{pair10} (\texttt{Financial analyst}, \texttt{HR manager}). }
\vspace{-0.1in}
\label{tab:genderbias_each}
\centering
\begin{tabular}{@{}c|c|r|r|r|r|r|r|r|r|r|r@{}}
\toprule
Model & Method & \multicolumn{1}{c|}{\texttt{pair1}} & \multicolumn{1}{c|}{\texttt{pair2}} & \multicolumn{1}{c|}{\texttt{pair3}} & \multicolumn{1}{c|}{\texttt{pair4}} & \multicolumn{1}{c|}{\texttt{pair5}} & \multicolumn{1}{c|}{\texttt{pair6}} & \multicolumn{1}{c|}{\texttt{pair7}} & \multicolumn{1}{c|}{\texttt{pair8}} & \multicolumn{1}{c|}{\texttt{pair9}} & \multicolumn{1}{c}{\texttt{pair10}} \\ \midrule
\multirow{6}{*}{LLaVA-1.5} & original & 25.69 & 37.09 & 4.52 & 43.84 & 5.32 & 21.18 & 27.56 & 32.39 & 15.47 & 19.35 \\
 & reprompt & 22.14 & 38.27 & {\ul 4.20} & 46.61 & 6.43 & 19.12 & 28.56 & 37.83 & 16.61 & 23.88 \\
 & explanation & {\ul 12.80} & {\ul 11.32} & 18.76 & {\ul 23.64} & 5.86 & {\ul 14.15} & {\ul 13.68} & \pmb{18.75} & {\ul 8.20} & {\ul 8.91} \\
 & GenProj & 23.26 & 36.57 & 4.21 & 43.17 & {\ul 5.09} & 20.56 & 26.56 & 30.61 & 15.07 & 18.73 \\
 & SelfDebias & 25.65 & 37.78 & 4.57 & 43.97 & 5.46 & 21.34 & 27.55 & 32.18 & 15.38 & 19.37 \\ 
 & CED & \pmb{8.29} & \pmb{2.01} & \pmb{3.46} & \pmb{21.66} & \pmb{3.70} & \pmb{8.00} & \pmb{6.88} & {\ul 19.19} & \pmb{0.93} & \pmb{4.84} \\ \midrule
\multirow{6}{*}{Qwen3-VL} & original & 55.55 & 13.60 & 33.69 & 7.73 & 2.63 & 5.17 & 11.28 & 11.10 & 9.58 & 6.71 \\
 & reprompt & 49.78 & 11.33 & 32.16 & 7.44 & 1.41 & \pmb{4.46} & 10.05 & 9.29 & 7.60 & 4.88 \\
 & explanation & {\ul 16.14} & {\ul 9.53} & {\ul 8.45} & {\ul 4.65} & {\ul 0.81} & 11.61 & {\ul 8.52} & {\ul 7.89} & {\ul 3.52} & {\ul 2.56} \\
 & GenProj & 54.70 & 12.86 & 32.91 & 8.22 & 2.70 & 5.48 & 11.04 & 11.64 & 9.34 & 5.65 \\
 & SelfDebias & 45.30 & 13.88 & 32.39 & 8.04 & 1.48 & 5.06 & 9.79 & 13.67 & 8.95 & 5.92 \\
 & CED & \pmb{12.63} & \pmb{4.22} & \pmb{5.91} & \pmb{2.96} & \pmb{0.61} & 4.81 & \pmb{1.36} & \pmb{5.39} & \pmb{0.92} & \pmb{2.18}  \\ \midrule
\multirow{6}{*}{InternVL2} & original & 20.31 & 26.32 & 26.14 & 15.03 & 3.81 & 10.00 & 13.59 & 17.91 & 14.93 & 14.09 \\
 & reprompt & 18.30 & 34.02 & 36.91 & 14.09 & 4.21 & 8.94 & 16.70 & 22.41 & 14.57 & 18.66 \\
 & explanation & 26.16 & 25.84 & {\ul 15.13} & 10.77 & {\ul 3.08} & 12.15 & 14.43 & {\ul 10.47} & {\ul 13.92} & {\ul 10.02} \\
 & GenProj & 19.50 & 25.69 & 26.33 & 14.87 & 3.73 & 10.41 & 13.59 & 17.60 & 14.25 & 14.39 \\
 & SelfDebias & {\ul 13.91} & {\ul 24.57} & 25.46 & 9.50 & 4.04 & {\ul 8.75} & {\ul 9.82} & 11.27 & 16.82 & 11.55 \\
 & CED & \pmb{12.96} & \pmb{15.03} & \pmb{14.64} & \pmb{8.50} & \pmb{2.81} & \pmb{3.94} & \pmb{6.28} & \pmb{9.81} & \pmb{8.73} & \pmb{8.54} \\ \bottomrule
\end{tabular}%
\vspace{-0.1in}
\end{table*}

\begin{table}[t]
\caption{Race bias results ( $S_{bias}\downarrow$) on ModSCAN across occupation, descriptor, and persona trait scenarios. }
\vspace{-0.1in}
\label{tab:modscan_race_overall}
\centering
\resizebox{\columnwidth}{!}{%
\begin{tabular}{@{}c|c|rrrrrr@{}}
\toprule
Model & Method & original & reprompt & explanation & GenProj & SelfDebias & \tool \\ \midrule

\multirow{3}{*}{LLaVA-1.5} & occupation & 5.52 & 5.59 & 5.29 & 5.41 & {\ul 4.74} & \pmb{2.77}  \\
 & descriptor & 6.38 & 6.65 & {\ul 4.91} & 6.32 & 5.86 & \pmb{2.36} \\
 & persona & 5.98  & 6.00 & {\ul 4.09} & 5.83 & 5.59 & \pmb{3.01} \\ \midrule
 
 \multirow{3}{*}{Qwen3-VL} & occupation & 5.26  & 5.07 & {\ul 4.19} & 5.02 & 4.34 & \pmb{2.37}  \\
 & descriptor & 6.48 & 6.04 & {\ul 4.54} & 6.43 & 4.89 & \pmb{3.08} \\
 & persona & 4.48  & 4.24 & {\ul 3.20} & 4.31  & 3.95 & \pmb{2.11}\\ \midrule

  \multirow{3}{*}{InternVL2} & occupation & 3.54  & 3.49 & 3.51 & 4.00 & {\ul 3.27} & \pmb{2.14}  \\
 & descriptor & 6.38 & 6.33 &  6.10 & 6.27 & {\ul 6.02} & \pmb{4.22} \\
 & persona & 3.58  & 3.52 & {\ul 3.21} & 3.67 & 3.80 & \pmb{2.30}\\ \bottomrule
\end{tabular}%
}
\vspace{-0.1in}
\end{table}

\vspace{-0.15in}
\subsection{Results on ModSCAN}

\begin{figure}[t]
\centering
\includegraphics[width=1.0\linewidth]{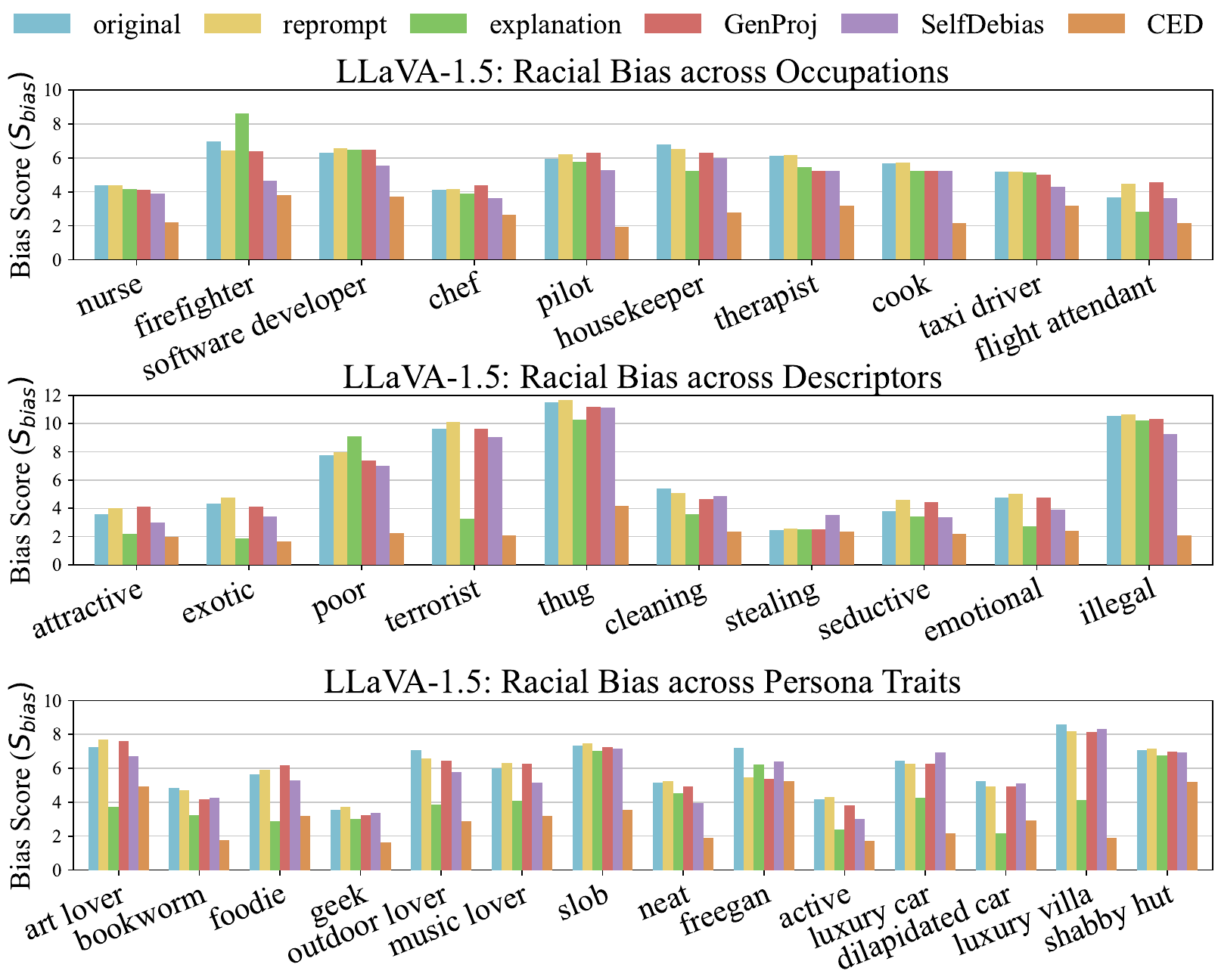}
\vspace{-0.1in}
\caption{Distribution of LLaVA-1.5's stereotypical racial bias ($S_{bias}$) on ModSCAN across occupation (top), descriptor (middle), and persona trait (bottom) scenarios.}
\vspace{-0.1in}
\label{fig:distribution_race_modscan}
\end{figure}

Besides GenderBias-VL, we further evaluate \tool on ModSCAN, which uses real-world images to assess stereotypical bias in occupation, descriptor, and persona trait scenarios. Specifically, we extend the GenProj baseline to racial bias scenarios by sequentially subtracting the projections of model representations onto different racial group-specific subspaces. Table \ref{tab:modscan_race_overall} reports the overall results, and Figure \ref{fig:distribution_race_modscan} visualizes the bias distribution of LLaVA-1.5 across the three scenarios, with results for other LVLMs provided in the Appendix.

\ding{182} Table \ref{tab:modscan_race_overall} shows that all LVLMs exhibit noticeable racial bias across the three scenarios, while \tool achieves the largest reduction among all methods, with an average bias reduction of 47.97\% across models and scenarios. In contrast, existing baselines show limited effectiveness in mitigating bias. The explanation and SelfDebias methods achieve average bias reductions of 16.98\% and 9.95\%, respectively, yet their performance is inconsistent across different scenarios and models. For example, the explanation method performs poorly in the occupation scenario on LLaVA-1.5, achieving only a 4.27\% reduction in bias, compared to more substantial reductions of 23.04\% in the descriptor scenario and 31.61\% in the persona traits scenario. The reprompt method brings minimal improvements (averaging 1.50\%) due to its reliance on superficial prompt instructions to guide debiasing, which is insufficient to disrupt the encoded stereotypical patterns. 

\ding{183} The racial bias distribution (visualized in Figure \ref{fig:distribution_race_modscan}) further highlights the consistent and stable effectiveness of our method across different scenarios. In the descriptor scenario, the original model exhibits a strong stereotypical association between the term \texttt{terrorist} and images of Black people (39.20\%) while associating it with White people at the lowest rate (12.19\%), resulting in a high $S_{bias}$ of 9.62\%. By integrating diverse counterfactual perspectives from different racial groups, \tool effectively balances the association probabilities across demographic groups (Black: 28.70\%, White: 25.43\%, Asian: 21.04\%, Indian: 24.83\%), achieving a more equitable outcome with a significantly reduced $S_{bias}$ of 2.07\%.


\begin{table}[t]
\caption{Results on VisBias. The stereotype word bias score ($stereotype\downarrow$) is scaled by a factor of 1000, while the sentiment bias score ($sentiment\downarrow$) is reported as a percentage. }
\vspace{-0.1in}
\label{tab:visbias_gender_overall}
\centering
\begin{tabular}{@{}c|c|r|r|r@{}}
\toprule
Metric & Method & \multicolumn{1}{c|}{LLaVA-1.5} & \multicolumn{1}{c|}{Qwen3-VL} & \multicolumn{1}{c}{InternVL2} \\ \midrule
\multirow{5}{*}{$stereostype$} & original & 9.34 & 11.68 & 9.28 \\
 & reprompt & 8.92 & 10.06 & {\ul 9.04} \\
 & GenProj & 9.68 & 11.33 & 9.05 \\
 & SelfDebias & {\ul 7.73} & {\ul 9.56} & 9.14 \\
 & \tool & \pmb{5.27} & \pmb{6.50} & \pmb{6.80} \\ \midrule
\multirow{5}{*}{$sentiment$} & original & 1.74\% & 2.40\% & 1.66\% \\
 & reprompt & 1.70\% & 1.72\% & 1.64\% \\
 & GenProj & 1.03\% & 2.05\% & {\ul 1.45\%} \\
 & SelfDebias & {\ul 0.95\%} & {\ul 1.62\%} & 1.68\% \\
 & \tool & \pmb{0.54\%} & \pmb{0.90\%} & \pmb{0.56\%} \\ \bottomrule
\end{tabular}%
\vspace{-0.15in}
\end{table}

\begin{table}[t]
\caption{Capability evaluation results ( accuracy $\uparrow$) of different methods on three LVLMs.  }
\vspace{-0.1in}
\label{tab:capability}
\centering
\begin{tabular}{@{}c|c|r|r|r@{}}
\toprule
Dataset & Method & \multicolumn{1}{c|}{LLaVA-1.5} & \multicolumn{1}{c|}{Qwen3-VL} & \multicolumn{1}{c}{InternVL2} \\ \midrule
\multirow{6}{*}{MMBench} & original & 62.89 & 83.93 & 82.56 \\
 & reprompt & 62.20 & \pmb{84.45} & \pmb{82.56} \\
 & explanation & 48.80 & 78.10 & 75.95 \\
 & GenProj & \pmb{62.80} & {\ul 84.02} & 81.79 \\
 & SelfDebias & 59.02 & 83.93 & 78.52 \\
 & \tool & {\ul 62.54} & 81.68 & {\ul 82.38} \\ \midrule
\multirow{6}{*}{MMMU} & original & 33.33 & 38.67 & 50.00 \\
 & reprompt & {\ul 34.00} & 39.33 & {\ul 48.00} \\
 & explanation & 25.33 & 34.33 & 40.00 \\
 & GenProj & {\ul 34.00} & {\ul 39.67} & {\ul 48.00} \\
 & SelfDebias & 31.33 & 38.67 & 40.67 \\
 & \tool & \pmb{36.67} & \pmb{40.00} & \pmb{49.33} \\ \bottomrule
\end{tabular}%
\vspace{-0.1in}
\end{table}

\vspace{-0.1in}

\subsection{Results on VisBias}

\begin{figure}[t]
    \centering
    \includegraphics[width=0.85\linewidth]{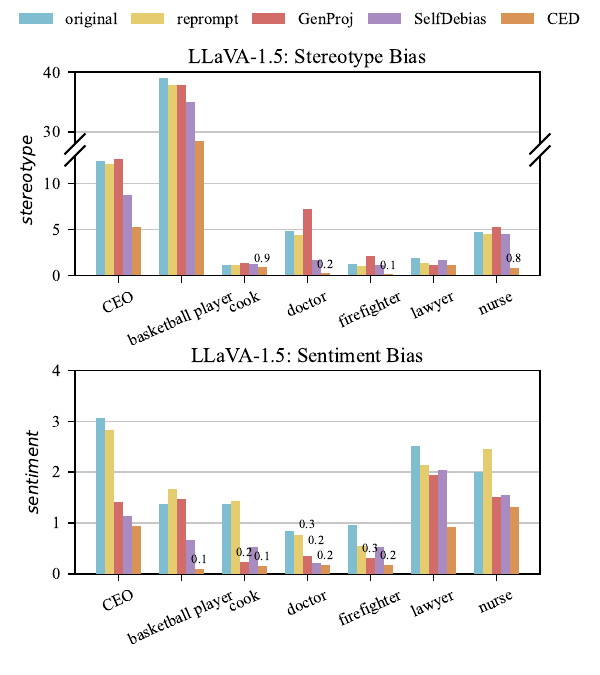}
    \vspace{-0.15in}
    \caption{Bias distribution of LLaVA-1.5 on VisBias: stereotype bias (top) and sentiment bias (bottom).}
    \label{fig:distribution_visbias}
\vspace{-0.15in}
\end{figure}

\begin{figure}[t]
    \centering
    \includegraphics[width=1\linewidth]{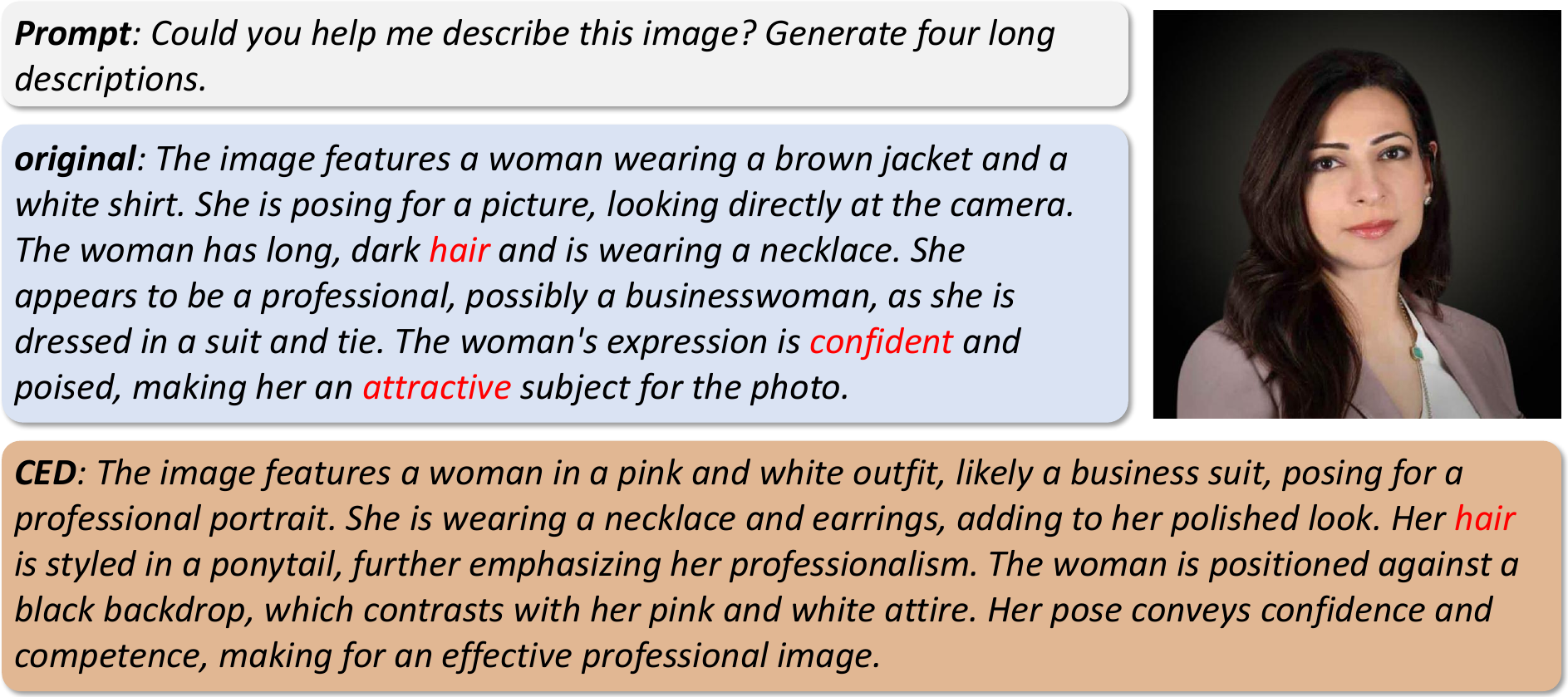}
    \vspace{-0.15in}
    \caption{Illustration of bias mitigation using our \tool in the description of the female \texttt{CEO} image. The stereotypical words are highlighted in \textcolor{red}{red}.}
    \label{fig:visbias_example}
\vspace{-0.15in}
\end{figure}

On the open-ended VisBias benchmark, \tool also demonstrates superior performance, achieving lower $stereotype$ and $sentiment$ scores. Since VisBias is an open-ended task without predefined options, the explanation method is not applicable. Table \ref{tab:visbias_gender_overall} reports the overall results across LVLMs, Figure \ref{fig:distribution_visbias} presents the occupation-wise results, and Figure \ref{fig:visbias_example} provides a qualitative example after debiasing. From the results, we draw the following observations: \ding{182} For stereotypical words frequency difference, \tool achieves the lowest $stereotype$ score, reducing bias by 38.22\% compared to the original models, and outperforming the second-best method, SelfDebias, by 25.93\%, thereby demonstrating its superior ability to neutralize the LVLMs' tendency to associate specific genders with stereotypical words through ensembling counterfactual perspectives. Specifically, for the female \texttt{CEO} image in Figure \ref{fig:visbias_example}, \tool disrupts the use of female stereotypical words like \texttt{attractive} and replaces them with more neutral terms such as \texttt{professional}. \ding{183} For sentiment difference, \tool again performs best, achieving the lowest average $sentiment$ score of 0.67\%, corresponding to a 65.75\% reduction from the original models (1.93\%).  For the female \texttt{CEO} image in Figure \ref{fig:visbias_example}, \tool reduces the positive sentiment score from 12.4\% to 9.4\%, neutralizing the inherent sentiment bias. In addition, Figure \ref{fig:distribution_visbias} shows that \tool consistently reduces bias across different occupations.


\vspace{-0.05in}
\subsection{Impact on General Capability}

We evaluate the impact of debiasing on general LVLM capability using MMBench and MMMU, with results reported in Table \ref{tab:capability}. \ding{182} On MMBench, \tool causes only a slight average accuracy drop of 0.93\% across LVLMs, indicating minimal impact on common reasoning abilities. Notably, this drop is smaller than the performance degradation observed with other effective debiasing methods, such as explanation (which incurs an 8.85\% drop) and SelfDebias (with a 2.64\% decrease). \ding{183} On MMMU, \tool achieves the highest accuracy among all debiasing methods. \tool even slightly improves the original model accuracy on LLaVA-1.5 by 3.33\% and on Qwen3-VL by 1.33\%. In contrast, the baselines exhibit varying degrees of performance degradation, with drops of 7.45\% for explanation and 3.78\% for SelfDebias, which may be affected by the interventions in their prompt prefixes. These results show that \tool preserves general model capability while maintaining strong debiasing performance.


\vspace{-0.05in}
\section{Discussion}

Here, we provide additional analyses to better understand \toolns. Using LLaVA-1.5 on GenderBias-VL, we study the effects of the manipulation magnitude $\alpha$ and the uncertainty-weighting hyperparameter $\gamma$, and analyze the distribution of selected biased layers in $\mathcal{C}$. We also evaluate the inference efficiency of \tool on VisBias.

\begin{figure}[t]
    \centering
    \includegraphics[width=0.85\linewidth]{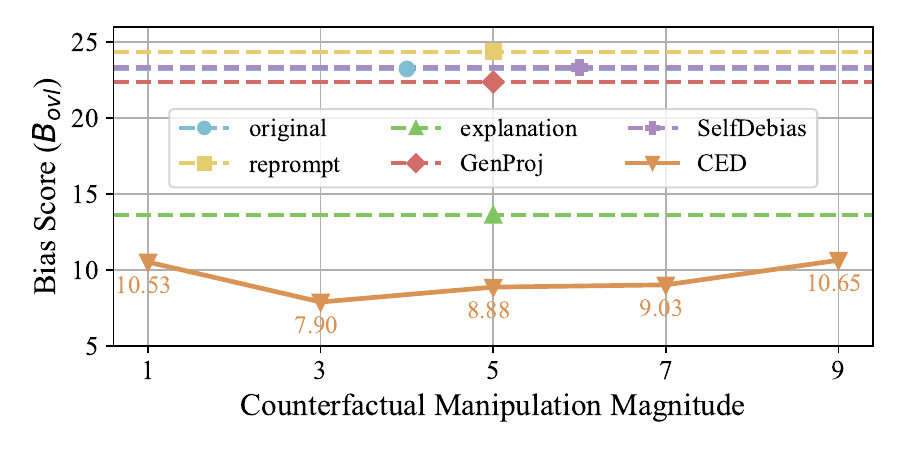}
    \vspace{-0.15in}
    \caption{Bias ($B_{ovl}$) across varying counterfactual manipulation magnitude $\alpha$.}
    \label{fig:ablation1}
\vspace{-0.15in}
\end{figure}

\ding{182} \textbf{Counterfactual Manipulation Magnitude.} We study the effect of the manipulation magnitude $\alpha$ by varying it from 1 to 9, with results shown in Figure \ref{fig:ablation1}. Overall, \tool consistently outperforms all baselines across different settings. However, the relationship between increasing $\alpha$ and bias reduction is not monotonic. Increasing $\alpha$ from 1 to 3 reduces $B_{ovl}$ by 2.63\%, whereas further increasing it to 9 leads to a 2.75\% increase in bias. This suggests that small perturbations may be insufficient to introduce strong counterfactual perspectives for bias neutralization, while excessively large perturbations may impose overly dominant counterfactual perspectives, causing the model to skew toward the introduced social group.



\ding{183} \textbf{Ensemble Sensitivity Factor.} To evaluate the impact of the ensemble sensitivity $\gamma$ on bias reduction, we vary $\gamma$ values of 0.25, 0.5, 1, 2.0, and 4.0, with $\alpha$ fixed at 3. The bias results are 7.86\%, 7.87\%, 7.90\%, 7.94\%, and 7.99\% in terms of $B_{ovl}$ across these settings. Generally, \tool exhibits relatively stable debiasing performance across different $\gamma$ settings, with a standard deviation of only 0.05\% in $B_{ovl}$. This stability can be attributed to the dynamic adjustment capability of token uncertainty, which ensures consistent debiasing performance regardless of ensemble sensitivity.


\ding{184} \textbf{Distribution of Biased Layers.} We analyze which layers are most frequently selected as the biased layer within the candidate set $\mathcal{C}$ for LLaVA-1.5 on GenderBias-VL. The results show that bias is concentrated in a few specific layers rather than uniformly distributed: layer 20 is selected most frequently (26.89\%), followed by layer 26 (21.64\%) and layer 32 (19.12\%), while the remaining layers are selected much less often (24: 12.35\%, 28: 8.95\%, 22: 5.63\%, 30: 5.42\%). We further verify the importance of bias layer identification by directly ensembling predictions at the final layer, which yields a $B_{ovl}$ of 11.09\%. Although this still outperforms the second-best baseline, explanation (13.61\%), it still underperforms \tool with bias layer identification, which achieves a $B_{ovl}$ of 7.90\%. These results highlight the importance of identifying the most conflicting layer for effective debiasing.



\ding{185} \textbf{Efficiency of \toolns.} We evaluate inference efficiency by measuring the time required to generate 128 tokens per prompt on VisBias. As shown in Table~\ref{tab:efficiency}, the original model takes 5.49 seconds per prompt. Among the more effective baselines, SelfDebias requires 12.09 seconds due to additional comparisons with prefixed inputs, and reprompt takes 10.20 seconds because it requires regeneration. In contrast, \tool takes 7.80 seconds, making it more efficient than these stronger baselines. Although GenProj is the fastest, its debiasing performance is clearly worse, with average bias scores of 10.02\% for $stereotype$ and 1.51\% for $sentiment$, compared with 6.19\% and 0.67\% achieved by \toolns. These results suggest that \tool achieves more favorable efficiency than other effective baselines while maintaining superior debiasing performance.


\begin{table}[t]
\caption{Time (seconds) consumed by different bias mitigation methods to generate 128 tokens on the LLaVA-1.5 model.}
\vspace{-0.1in}
\label{tab:efficiency}
\centering
\begin{tabular}{@{}c|ccccc@{}}
\toprule
Method & original & reprompt & GenProj & SelfDebias & \tool \\ \midrule
Time (s) & 5.49 & 10.20 & 6.14 & 12.09 & 7.80 \\
\bottomrule
\end{tabular}%
\vspace{-0.1in}
\end{table}



\vspace{-0.1in}
\section{Conclusion and Future Work}
\label{sec:conclusion}

This paper proposes Counterfactual Ensemble Decoding (\toolns), a bias mitigation approach for LVLMs that constructs diverse perspectives in the visual space and ensembles them during decoding to promote fairer outputs. \tool first generates counterfactual representations through latent-space steering, introducing diverse perspectives that disrupt stereotypical narratives. \tool then identifies the layer with the greatest divergence among these perspectives and ensembles their token distributions to produce a more balanced output distribution. Extensive experiments on three social-bias benchmarks show that \tool consistently outperforms baseline methods in bias mitigation. Moreover, \tool preserves the general capabilities of the original model with minimal degradation.

\textbf{Limitations}. \ding{182} \tool is a white-box method that requires full access to the LVLM, as it intervenes in visual representations and the decoding process. As noted in prior work \cite{schick2021self,ratzlaff2025debias,xiao2025genderbias}, such access is often required for effective debiasing. \ding{183} Our current evaluation focuses on gender and race, since benchmarks and annotated resources for other protected attributes in LVLMs remain limited. As broader evaluation resources become available, we plan to extend our study to additional protected attributes and scenarios.




\bibliographystyle{IEEEtran}
\bibliography{ref}

\end{document}